\documentclass[conference,preprint]{IEEEtran}  

\usepackage{graphics} 
\usepackage{epsfig} 
\usepackage{amsmath} 
\usepackage{amssymb}  
\usepackage{bm}
\usepackage{subcaption}
\usepackage{booktabs}
\usepackage{hyperref}

\title{\LARGE \bf
 An Online Optimization-Based Trajectory Planning Approach for Cooperative Landing Tasks
}

\author{
    \IEEEauthorblockN{Jingshan Chen\IEEEauthorrefmark{1}, Lihan Xu\IEEEauthorrefmark{1}, Henrik Ebel\IEEEauthorrefmark{2}, and Peter Eberhard\IEEEauthorrefmark{1}}
    \IEEEauthorblockA{\IEEEauthorrefmark{1}Institute of Engineering and Computational Mechanics, University of Stuttgart, Pfaffenwaldring 9, 70569 Stuttgart, Germany\\
    Emails: \{jingshan.chen, peter.eberhard\}@itm.uni-stuttgart.de, lihan.xu16@outlook.com}
    \IEEEauthorblockA{\IEEEauthorrefmark{2}Department of Mechanical Engineering, LUT University, Yliopistonkatu 34, 53850 Lappeenranta, Finland\\
    Email: henrik.ebel@lut.fi}
}

\begin{document}

\maketitle
\thispagestyle{empty}
\pagestyle{empty}

\begin{abstract}

        This paper presents a real-time trajectory planning scheme for a heterogeneous multi-robot system (consisting of a quadrotor and a ground mobile robot) for a cooperative landing task, where the landing position, landing time, and coordination between the robots are determined autonomously under the consideration of feasibility and user specifications. The proposed framework leverages the potential of the complementarity constraint as a decision-maker and an indicator for diverse cooperative tasks and extends it to the collaborative landing scenario. 
        In a potential application of the proposed methodology, a ground mobile robot may serve as a mobile charging station and coordinates in real-time with a quadrotor to be charged, facilitating a safe and efficient rendezvous and landing.
        We verified the generated trajectories in simulation and real-world applications, demonstrating the real-time capabilities of the proposed landing planning framework.  

\end{abstract}

\section{INTRODUCTION}

Heterogeneous multi-robot collaboration combines the advantages of different robot domains, providing wide-area coverage and high environmental adaptability. This cross-domain cooperation, where aerial and ground robots are usually involved, has been developed and utilized in various applications, such as search and rescue, logistics and delivery, and infrastructure inspection~\cite{Munasinghe2024}.

In this work, we focus on the coordinated landing of a quadrotor on a ground mobile robot. In a potential application, when the quadrotor's power becomes insufficient, a ground mobile robot, serving as a mobile charging station, may cooperate with the quadrotor to facilitate a safe and efficient landing for direct charging and later re-launch. This cooperative landing differs from the unilateral landing, where only one side of the air-ground robot system adjusts its response while the other side continues following a predefined path. 

Regarding air-ground collaboration, most works are unilateral cooperations of the unmanned aerial vehicle (UAV), i.e., the current position and attitudes of the unmanned ground vehicle (UGV) are known, either detected by sensors or predicted by algorithms. Then, the UAV should, based on known information, autonomously land on the ground mobile robot. The landing trajectories can be derived through different algorithms, such as deep reinforcement learning~\cite{RodriguezRamos2018}, a vision-based and pure pursuit guidance~\cite{Gautam2022}, or other motion planning methods such as rapidly-exploring random trees (RRT)~as mentioned in~\cite{Nurimbetov2017} and~\cite{Sinnemann2022}.
Alternatively, in~\cite{Si2024}, a landing system based on the ground vehicle is developed to locate, track, and coordinate the UAV for landing. Nevertheless, when focusing solely on the landing process, these unilateral procedures often prove inefficient due to the lack of interactive coordination between the robots, making it challenging to achieve higher task completion rates and complete automation in logistics. 

In recent years, some research has explored the cooperation of UAVs and UGVs for an interactive landing. For instance,~\cite{Rabelo2021} employs formation control to first achieve a predefined distance between the UAV and the ground mobile robot, enabling vertical landing. However, this approach involves multiple landing steps, making the landing time-consuming, and it only considers kinematic models. This is a critical limitation when a complete dynamic model of the UAV is required for agile dynamic and precise landing. Meanwhile,~\cite{Hebisch2021} adopts a model predictive control scheme consisting of vertical and horizontal trajectory planners. The planned trajectories are separately transmitted to the robots' trajectory tracking controllers. 
Whereas this approach is efficient in terms of computation time, it relies exclusively on a linear model predictive controller for trajectory planning, which leads to a loss of model accuracy and input constraint approximation issues. 
Additionally, the generated trajectories have not yet been experimentally validated for their feasibility on actual hardware.

Considering system dynamics and constraints is crucial for generating feasible trajectories. Typically, feasible trajectory planning follows a two-steps process. Firstly, a path consisting of a few predetermined waypoints is established based on some task-specific requirements. For instance, in drone racing, specific waypoints must be traversed, or a route may be roughly designed to prioritize obstacle avoidance. 
Classically, only in a subsequent step, this path is time-parameterized to create a feasible trajectory. However, if the system dynamics and constraints are ignored in the first step, a dynamically feasible trajectory may not be derived from this path or may not even exist~\cite{Tang2020}. Even if it can be derived, a result of such a two-stage process will generally be worse than one from a joint, one-step planning process.
Another key challenge in planning interactive landing is determining where, when, and how both robots should coordinate for the landing. We aim to avoid manual intervention in the decision-making process. Therefore, based on the investigation discussed in our previous works~\cite{Luo2023} and~\cite{Chen2024}, the complementarity constraint proves to be a suitable method for autonomous decision-making and can be seamlessly integrated into optimization problems, allowing for the inclusion of additional requirements within the same framework. This approach is effective not only for the decision-making of multibody systems but also for multi-robot systems, enabling coordinated operations.

The novel contributions of this work can be summarized as follows:

\begin{itemize}
  \item a one-step trajectory planning framework for the cooperative and coordinated landing of heterogeneous multi-robot systems
  \item consideration of system dynamics and user requirements while simultaneously addressing time-allocation
  \item autonomous decision-making through integrated complementarity constraints
  \item time optimization of tasks for efficiency and flexible time scaling by treating each integration time interval as an optimization variable
  \item online planning capability tested both in simulation and real-world applications
\end{itemize}

This paper consolidates the main ideas of our previous paper~\cite{Luo2023} and~\cite{Chen2024}, integrating autonomous decision-making and feasibility in trajectory planning, and extends these concepts to a coordinated landing scenario with a significant improvement in computation time. 
The latter is crucial for online replanning, which is critical for real-world robustness. For instance, in the recent paper~\cite{Antal2025}, the authors recognize that our previous approach is ``powerful'' and ``without the inherent suboptimality'' of other approaches, but they use the slowness of our previous approach to come up with another suboptimal approach. Instead, in this paper, we show how to make our approach work in real-time, orders of magnitude faster than before.

The investigated multi-robot system is introduced in Section~\ref{sec:modeling}, followed by the design of the landing trajectory planning framework in Section~\ref{sec:methodology}. Numerical and experimental results are presented in Sections~\ref{sec:numerical_evaluation} and~\ref{sec:hardware_experiment}, and finally, Section~\ref{sec:conclusion} summarizes our work and provides an outlook on future work.

\section{Modeling}
\label{sec:modeling}

\begin{figure}
  \centering
  \begin{minipage}{0.45\columnwidth}
      \centering
      \includegraphics[width=0.8\textwidth]{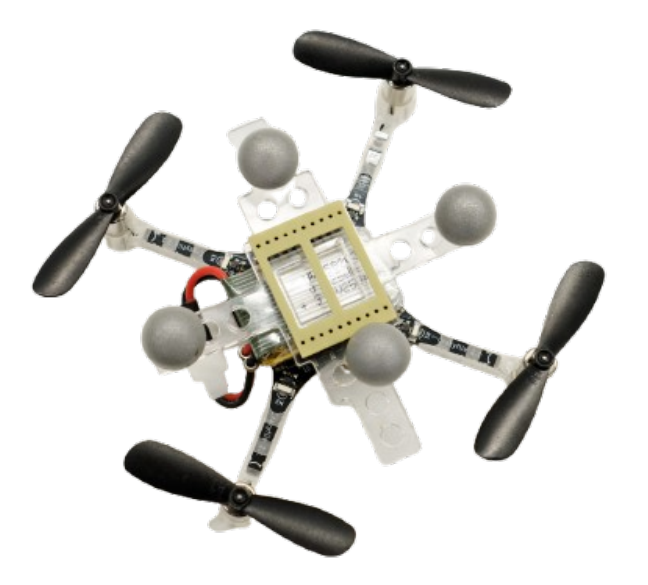}
      \caption{Nano quadrotor \textit{Crazyflie}~\cite{Giernacki2017}}
      \label{fig:quadrotor}
  \end{minipage}\hfill
  \begin{minipage}{0.45\columnwidth}
      \centering
      \includegraphics[width=\textwidth]{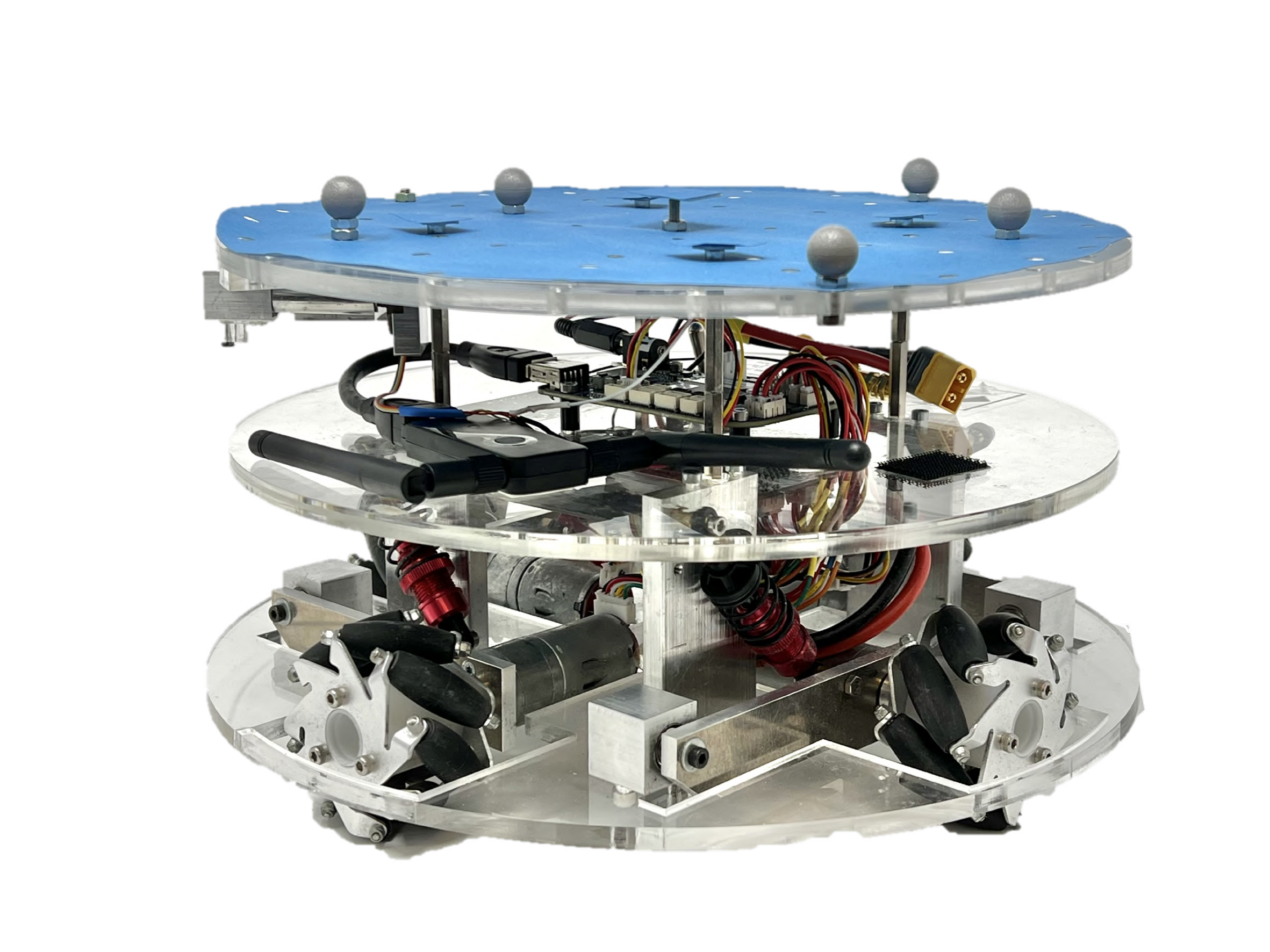}
      \caption{Custom-built mobile robot \textit{HERA}~\cite{Ebel2021}}
      \label{fig:omni}
  \end{minipage}
\end{figure}

This section covers the modeling of both the quadrotor and the omnidirectional mobile robot, shown in Figs.~\ref{fig:quadrotor} and ~\ref{fig:omni}, which are the agents of the heterogeneous multi-robot system investigated in our cooperative landing scenario.

\subsection{Dynamics of the Quadrotor}
We define the state of the quadrotor as $\bm{x}_\text{q} := \left[\bm{p}_\text{q}^\top \ \bm{\xi}^\top \ \bm{v}^\top \ \bm{\omega}_\text{B}^\top \right]^\top \in \mathbb{R}^{12} $,
where $\bm{p}_\text{q}:= \left[x\ y\ z\right]^\top$ is the position in the inertial frame, and $\bm{v}=\dot{\bm{p}}_\text{q}:=\left[v_x \ v_y \ v_z \right]^\top$ denotes the corresponding velocity. The vector $\bm{\xi}:=\left[\phi \ \theta \ \psi\right]^\top$ contains the Euler angles, while $\bm{\omega}_\text{B}$ represents the angular velocity of the quadrotor. 
Note that quantities expressed in the body-fixed coordinate frame are denoted with the subscript $(\cdot)_\text{B}$.  
The input of the quadrotor is defined as 
$\bm{u}_\text{q} = \left[f_{\text{1}} \ f_{\text{2}} \ f_{\text{3}} \ f_{\text{4}}\right]^\top$, where $f_{i}$ represents the force generated by the $i$-th propeller. The collective thrust along the $z$-axis of the quadrotor's body-fixed frame is given by $f_\text{thrust} = \sum_{n = 1}^{4}f_{i}$. 
The torque acting on the quadrotor in the body-fixed frame is then expressed as
\begin{equation*}
  \label{eq:torque_motors}
  \bm{\tau}_{\text{B}}=\begin{bmatrix}
    \frac{\sqrt{2}}{4}\ell (f_\text{2}+f_\text{3}-f_\text{1}-f_\text{4}) \vspace{1mm} \\ \frac{\sqrt{2}}{4}\ell (f_\text{2}+f_\text{4}-f_\text{1}-f_\text{3}) \vspace{1mm}\\ c_\tau (f_\text{3}+f_\text{4}-f_\text{1}-f_\text{2})
  \end{bmatrix},
\end{equation*}
where $\ell$ is the diagonal length of the quadrotor frame and $c_\tau$ is the torque coefficient~\cite{Quan2017}. 
The dynamics of the quadrotor can be derived as
\begin{equation}
  \label{eq:quad_system_dyanmics}
  \dot{\bm{x}}_\text{q} = f_\text{q}(\bm{x}_\text{q}, \bm{u}_\text{q})= \begin{bmatrix}
    \bm{v} \\
    \bm{T}(\bm{\xi})\bm{\omega}_\text{B} \\
    \bm{R}_\text{B}(\bm{\xi}) \left[0 \ 0 \ f_\text{thrust}\right]^\top - \left[0 \ 0 \ m_\text{q}g\right]^\top \\
    \bm{J}_\text{q}^{-1}(-\bm{\omega}_\text{B}\times\bm{J}_\text{q}\bm{\omega}_\text{B} + \bm{\tau}_\text{B})
  \end{bmatrix},
\end{equation}
where $m_\text{q}$ is the mass of the quadrotor, $\bm{J}_\text{q}=\text{diag}(J_{xx},J_{yy},J_{zz})$ is the quadrotor inertia matrix, $\bm{R}_\text{B}(\bm{\xi})$ is the rotation matrix from the body-fixed frame to the inertial frame, and $\bm{T}(\bm{\xi})$ is the transformation matrix mapping the angular velocity from the body-fixed frame to the inertial frame, defined as
\begin{equation*}
  \bm{T}(\bm{\xi}) = \left[\begin{array}{ccc}
      1 & \sin(\phi)\tan(\theta) & \cos(\phi)\tan(\theta) \\
      0 & \cos(\phi)             & -\sin(\phi)            \\
      0 & \frac{\sin(\phi)}{\cos(\theta)} & \frac{\cos(\phi)}{\cos(\theta)}
    \end{array}\right].
\end{equation*}

\subsection{Dynamics of the Ground Mobile Robot}

The modeling of mobile robots is based on our customized omnidirectional mobile \textit{HERA} robot, which can move freely in the plane in any direction. The state vector is defined as $\bm{x}_{\text{omni}} := [x_\text{omni}\ y_\text{omni}\ v_{x, \text{omni}}\ v_{y, \text{omni}}]^\top \in \mathbb{R}^{4}$.
The input vector is selected as $\bm{u}_\text{omni} := \left[f_\text{omni}\  \zeta\right]^\top \in \mathbb{R}^{2}$, where $f_\text{omni}$ is the propulsion force acting on the horizontal plane and $\zeta$ is the angle between $f_\text{omni}$ and the $x$-axis of the inertial frame of reference.
The dynamics of the ground mobile robot is represented as
\begin{equation}
  \label{eq:omni_dynamics}
      \dot{\bm{x}}_{\text{omni}}= f_\text{omni}(\bm{x}_{\text{omni}}, \bm{u}_{\text{omni}}) =
      \begin{bmatrix}
      v_{x, \text{omni}}\\
      v_{y, \text{omni}}\\
      f_\text{omni}\cos{(\zeta)}/m_{\text{omni}}\\
      f_\text{omni}\sin{(\zeta)}/m_{\text{omni}}
  \end{bmatrix},
  \end{equation}
  where $m_{\text{omni}}$ is the mass of the robot.

\section{Methodology}
\label{sec:methodology}
In this section, we introduce the main components of the optimization-based trajectory planning framework for the cooperative landing task, consisting of the complementarity constraints for autonomous decision-making and time optimal control. 

\subsection{Autonomous Decision-Making}

The complementarity constraint is based on the principle $$ab = 0,\ a \geq 0,\ b \geq 0.$$ This implies that at most one of the consisting components of the constraint can be positive, which allows for a straightforward physical interpretation. In the context of the cooperative landing, one component of the constraints is the landing index variable $\varepsilon \in \left[0, 1\right]$, whose value is autonomously determined by solving the optimization problem under the consideration of other constraints. The other component corresponds to the distance between the quadrotor and the landing platform $f_{\text{dis}}= \left\Vert \bm{p}_{\text{q}} - \bm{p}_{\text{omni}} \right\Vert$ using the Euclidean norm, where $\bm{p}_{\text{omni}}=\left[x_\text{omni}\ y_\text{omni}\ z_\text{omni}\right]^\top$ represents the position of the ground mobile robot with a constant height $ z_\text{omni}$. To address numerical intractability, the complementarity constraint is relaxed by introducing a relaxation variable $\nu \in \left[0, \nu_{\text{max}}\right]$. Then, the complementarity constraint for each time step $k \in \left\{0,\ldots,N-1\right\}$ can be formulated as
\begin{equation}
  \label{eq:complementarity}    
  \varepsilon_k \left(f_{\text{dis},k}-\nu_k \right) = 0.
\end{equation}
At the beginning of the landing process, the distance between the quadrotor and the mobile robot is positive, whereas the landing index variable $\varepsilon_k$ must be zero. Once the robots enter the suitable range for landing, considered by $\nu_\text{max}$, the landing index variable can be nonzero. In this context, the value of $\varepsilon_k$ serves as an indicator of the landing process: when it becomes to nonzero, the landing operation is considered completed.
However, by definition, the landing index variable $\varepsilon_k$ may remain zero even when the distance requirement for landing is satisfied. To ensure that the landing index variable changes its value at least at one time step, additional constraints are required, which are defined as
\begin{equation}
  \label{eq:additional_progress_condition}
  \begin{aligned}
    \kappa_{k+1} & = \kappa_{k} - \varepsilon_{k} , \hspace{2mm} k \in \{0, \ldots, N-1\}                          \\
    \text{with} & \hspace{2mm} \kappa_0 = 1\hspace{2mm}\text{and}\hspace{2mm} \kappa_N = 0.
  \end{aligned}
\end{equation}

\subsection{Time Optimal Control}
\label{sec:time_optimal_control}
To achieve an efficient landing process, the optimal control problem of searching for minimum total time and corresponding control trajectory is considered in landing trajectory generating. We adopt the local uniform grid approach described in~\cite{Roesmann2021}, where each integration interval is treated as an optimization variable $\Delta t_k$.
Notably, we do not impose the uniformity condition that requires $\Delta t_k = \Delta t_{k+1}$. This flexibility allows for adaptive time scaling and time allocation.

\subsection{Landing Trajectory Planning Scheme}
\label{sec:landing_trajectory_planning_scheme}
The cooperative landing planning framework can be formulated as a finite-horizon discrete optimal control problem
\begin{equation}
  \label{eq:final_optimization}
  \begin{aligned}
    \min_{\bm{x}^*} \hspace{3mm} & J_\text{d} \left(\Delta t_k, \bm{x}_{\text{q},k}, \bm{x}_{\text{omni},k}, \bm{u}_{\text{q},k}, \bm{u}_{\text{omni},k}, \kappa_k, \nu_{k} \right) 
    \\
    \text{s.t.} \hspace{3mm} & \bm{x}_{\text{q},0} = \bm{x}_{\text{q}}(0), \hspace{2mm} \bm{x}_{\text{omni},0} = \bm{x}_{\text{omni}}(0)
    \\
    & \bm{x}_{\text{q},k+1} = f_\text{RK}(\bm{x}_{\text{q},k}, \bm{u}_{\text{q},k}, \Delta t_k)
    \\
    & \bm{x}_{\text{omni},k+1} = f_\text{RK}(\bm{x}_{\text{omni},k}, \bm{u}_{\text{omni},k}, \Delta t_k)
    \\ 
    & \hspace{33mm} \forall  k\in \{0, \ldots, N-1\},
    \\
    & \bm{x}_{\text{q},\text{max}} \leq \bm{x}_{\text{q},k} \leq \bm{x}_{\text{q},\text{max}}
    \\
    & \bm{x}_{\text{omni},\text{max}} \leq \bm{x}_{\text{omni},k} \leq \bm{x}_{\text{omni},\text{max}}
    \\
    & \hspace{33mm} \forall  k\in \{0, \ldots, N\},
    \\
    & \bm{u}_{\text{q},\text{max}} \leq \bm{u}_{\text{q},k} \leq \bm{u}_{\text{q},\text{max}}
    \\
    & \bm{u}_{\text{omni},\text{max}} \leq \bm{u}_{\text{omni},k} \leq \bm{u}_{\text{omni},\text{max}}
    \\
    &  \hspace{33mm} \forall  k\in \{0, \ldots, N-1\},
    \\
    & \text{complementarity constraints~\eqref{eq:complementarity} and~\eqref{eq:additional_progress_condition}},
    \\
    & 0 \leq \varepsilon_{k} \leq 1 \hspace{17mm} \forall  k\in \{0, \ldots, N-1\}, 
    \\
    & 0 \leq \nu_{k} \leq \nu_\text{max} \hspace{13mm} \forall  k\in \{0, \ldots, N-1\}, 
    \\
    & \Delta t_\text{min} \leq \Delta t_k \leq \Delta t_\text{max} \hspace{2mm} \forall  k\in \{0, \ldots, N\}
  \end{aligned}
\end{equation}
with the optimizer $\bm{x}^*={\left[ \bm{x}_{0}^*, \ldots ,\bm{x}_{N}^*\right]}$,
where 
\begin{equation*}
  \label{eq:grasping_opt_variables_1}	
  \renewcommand{\arraystretch}{1}
	\bm{x}_k^* = \left\{ \begin{array}{ll}
		\displaystyle \left[\bm{x}_{\text{q},k}^*, \bm{x}_{\text{omni},k}^*, \bm{u}_{\text{q},k}^*, \bm{u}_{\text{omni},k}^*, \varepsilon_k^*, \kappa_{k}^*,\nu_k^*, \Delta t_k^*\right] \\
    \hspace{4cm} \text{for} \ \ k\in[0,N) 
    \\
		\displaystyle \left[\bm{x}_{\text{q},N}^*, \bm{x}_{\text{omni},N}^*,\kappa_{N}^*, \Delta t_N^* \right] 
    \\ 
    \hspace{4cm}\text{for}\ \ k=N.
	\end{array} \right.
\end{equation*}
The second-order Runge-Kutta method is utilized to solve the dynamics of the quadrotor and the omnidirectional robot, indicated by $f_\text{RK}$.
Moreover, we employ the nonlinear programming solver \textsc{FATROP}~\cite{Vanroye2023}, which achieves a significant reduction in computation time by exploiting the constrained optimal control problem's structure as demonstrated in Sec.~\ref{sec:computation_time}.

\section{Numerical Evaluation}
\label{sec:numerical_evaluation}

\subsection{Initialization Setup}
\label{sec:initialization_setup}
While appropriate initialization may improve optimization efficiency, improper initialization can lead to suboptimal solutions or increased solver time. 
Therefore, in the following, the optimization is initialized only with hover thrust for the quadrotor's input. 
The initial guesses for the other variables are set to zero in this work, to prevent interference in the landing decision-making process.

\subsection{Objective Function}
\label{sec:objective_function}

\begin{figure*}
  \centering
  \begin{minipage}{\columnwidth}
      \centering
      \includegraphics[width=\textwidth]{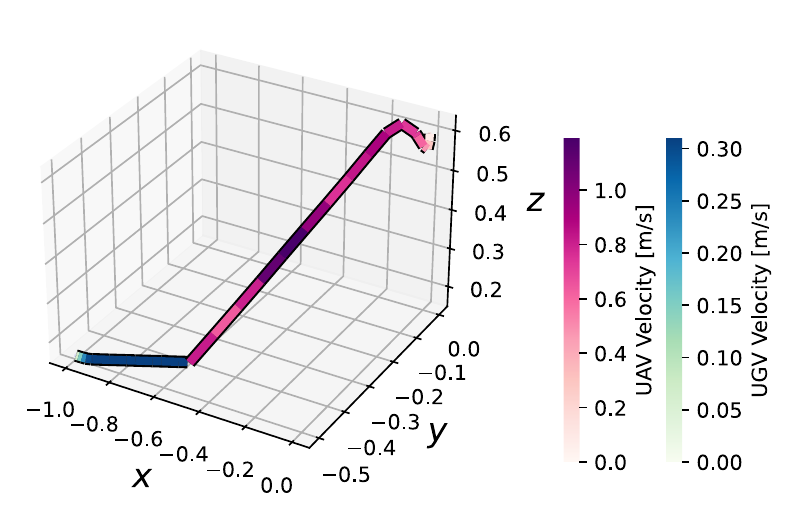}
      \subcaption{Generated landing trajectories with the objective function~\eqref{eq:objective_function}.}
      \label{fig:w5_0}
  \end{minipage}\hfill
  \begin{minipage}{\columnwidth}
      \centering
      \includegraphics[width=\textwidth]{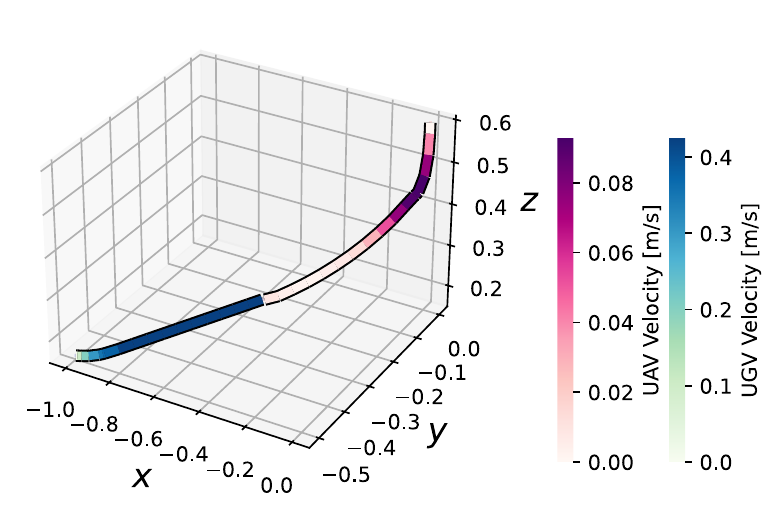}
      \subcaption{Generated landing trajectories considering the cost of quadrotor states in~\eqref{eq:objective_function} additionally.}
      \label{fig:w5_1}
  \end{minipage}
  \caption{Comparison of landing trajectories for different objective functions with the weighting value $w_3=1$ and $w_3=0$. Notationally, we use UAV to refer to the quadrotor and UGV to refer to the ground mobile robot. This annotation applies to all figures below.}
  \label{fig:uav_ugv_position_comparision}
\end{figure*}

Depending on user requirements, the objective function can be typically structured to account for total task time or control efforts. 
Our objective is to achieve an agile and cooperative landing process, which is why we do not assign specific importance to the control effort of any individual robot. Instead, the total time for the landing process is prioritized in the objective function.
Moreover, the formulation of the complementarity constraints allows the value of the landing index variable $\varepsilon$ to be positive at distinct time instants, which leads to an unreasonable landing performance, i.e., a sequence of landing, relaunching, and landing again. Additionally, as previously explored in~\cite{Chen2024}, without imposing constraints on the landing index variable, the transition of this variable from 0 to 1 (corresponding to $\kappa$ changing from 1 to 0) can not be assured at one single time instant, potentially leading to a suboptimal solution, where the landing index variable changes it value gradually over multiple time intervals, even when the distance requirement for landing is satisfied. To address those issues, we incorporate the regularization term of $\kappa$ into the objective function, which is defined as
\begin{equation}
  \label{eq:objective_function}
  \begin{aligned}
  J_\text{d} =\, & w_1\sum_{k=0}^{N}\Delta t_k + w_2\sum_{k=0}^{N}\kappa_k,
  \end{aligned}
\end{equation}
where the weighting factors are subsequently chosen as $w_1=20$ and $w_2=1$.

\begin{figure}
  \centering
  \includegraphics[width=0.8\columnwidth]{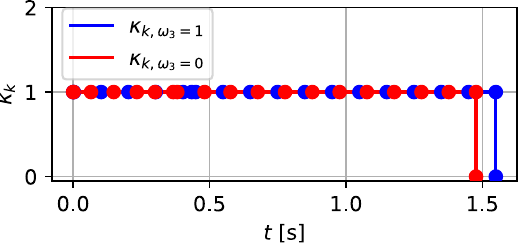}
  \caption{Comparison of the total time for different objective functions with the weighting value $w_3=1$ and $w_3=0$.}
  \label{fig:total_time_comparision}
\end{figure}

In Fig.~\ref{fig:w5_0}, we present the generated landing trajectories corresponding to the objective function~\eqref{eq:objective_function}. The quadrotor successfully executes the landing on the mobile robot. However, this occurs with a considerable velocity at the point of encounter. When the cost of the quadrotor states is considered in the objective function with an additional term $w_3\sum_{k=0}^N \left\Vert \bm{x}_{\text{q},k}\right\Vert^2$ and $w_3=1$, a less aggressive landing behaviour can be obtained as depicted in Fig.~\ref{fig:w5_1}.
Notably, the total time for both scenarios, represented by changes in the value of $\kappa$ over time in Fig.~\ref{fig:total_time_comparision}, is approximately 1.5\,s, with only slight differences observed.
Consequently, we include the cost of the quadrotor states within the objective function to facilitate a gentler landing process, thereby also enhancing the overall landing performance in hardware experiments.

\subsection{Computation Time}
\label{sec:computation_time}
It is important to note that the total number of time steps $N$ directly affects computational complexity, as the number of variables and constraints scales proportionally to it. To determine an appropriate value for $N$, we base our estimation on the relationship between the maximal velocity, the initial linear distance between the robots, and the maximum time interval
$$N_\text{est}= \frac{f_{\text{dis},t_0}}{\Delta t_\text{max}v_\text{max}}.$$ 
Due to our laboratory setup, we select $N$ typically to be within the range of $[20,30]$. 

Although our optimization problem is non-convex and includes nonlinear constraints, a feasible solution can still be achieved in approximately 0.1\,s using the \textsc{FATROP} solver, about two orders of magnitude faster than for the related problem in~\cite{Luo2023}. 
This allows for real-time replanning capabilities to address potential disturbances and uncertainties online. 

\begin{figure}[thpb]
  \centering
  \includegraphics[width=0.8\columnwidth]{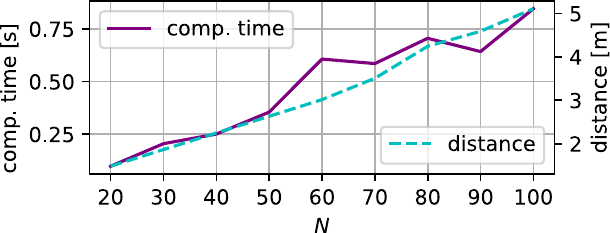}
  \caption{Comparison of computation time for different numbers of time steps $N$ corresponding to the initial linear distance between the robots in different simulations.} 
  \label{fig:compTime_comparision_N}
\end{figure}

To further examine the computational capacity, we conduct a series of simulations with varying numbers of time steps while maintaining the same setting. Given that the parameters and system constraints are fixed, our estimation indicates that the number of time steps increases with the linear distance between the agents, as illustrated by the blue dotted line in Fig.~\ref{fig:compTime_comparision_N}. The computation time, represented by the violet line, demonstrates a linear increase in relation to the number of time steps. Importantly, even for considerable distance landings with a high number of time steps, the computation time remains well under 1\,s, which is well acceptable for replanning during operation. Besides, the results of the time intervals do not exceed the maximum time interval $\Delta t_\text{max}$, enabling the generation of uniformly distributed waypoints with time allocation along the trajectory.

\section{Hardware Experiment}
\label{sec:hardware_experiment}

We now conduct hardware experiments to evaluate the real-world performance of our proposed trajectory planning framework for cooperative landing between a quadrotor and a mobile robot.

\subsection{Experimental Setup}
In our real-world experiments, we employ the \textit{Crazyflie~2.1} nano quadrotor developed by \textit{Bitcraze}, alongside a custom-built ground mobile robot \textit{HERA} from our institute to perform the cooperative landing task. Due to the limitation of the on-board processing unit, the trajectory planning and trajectory tracking controllers run on a portable laboratory laptop. The communication from the laptop to the quadrotor happens through the USB radio dongle \textit{Crazyradio~2.0}. In parallel, the communication to the mobile robot is established via the local wireless network using LCM (Lightweight Communications and Marshalling) messages~\cite{Huang2010}. 
Both robots are equipped with markers for the \textit{OptiTrack} motion capture system, which consists of six cameras operating at 240\,Hz to provide real-time measurement feedback. 
The values of the system parameters and the user-defined parameters are provided in Table~\ref{tab:parameters}.

\begin{table}[htp]
  \caption{Parameters for numerical simulations and hardware experiments}
  \label{tab:parameters}
  \centering
  \begin{tabular}{lll}
    \toprule
    Parameters & Value & Unit \\
    \midrule
    $m_\text{q}$ & 37.1 & g \\
    $m_\text{omni}$ & 3.2 & kg \\
    $\nu_\text{max}$ & 0.01 & m \\
    $\Delta t_\text{max}$ & 0.1 & s \\
    max./min. $\dot{\bm{p}_\text{q}}^\top$ & $\left[\pm0.5 \ \pm0.5 \ \pm0.5\right]$ & m/s \\
    max./min. $\dot{\bm{\xi}}^\top$ & $\left[\pm0.4 \ \pm0.4 \ \pm3.48\right]$ & rad/s \\
    max./min. $f_{i\text{-th}}$ & $1.5 f_\text{hover}$ $/$ $0.5 f_\text{hover}$ & N \\
    $f_\text{hover}$ & $0.25m_{\text{q}}g$ & N \\
    $\bm{J}_\text{q}$ & diag$\left(1.43, 1.43, 2.89\right)10^{-5}$ & kg\,m$^2$ \\
    $\ell$ & 92 & mm \\
    $c_\tau$ & 5.964 & mm \\
    max./min. $[\dot{x}_\text{omni}\ \dot{y}_\text{omni}]$ & $\left[\pm0.3 \ \pm0.3\right]$ & m/s \\
    max./min. $\bm{u}^\top_{\text{omni}}$ & $\left[\pm1.0 \ \pm\pi\right]$ & [N rad] \\
    \bottomrule
  \end{tabular}
\end{table}

\subsection{Trajectory Tracking}
\label{trajectory_tracking}
\begin{figure}[htpb]
  \centering
  \includegraphics[width=\columnwidth]{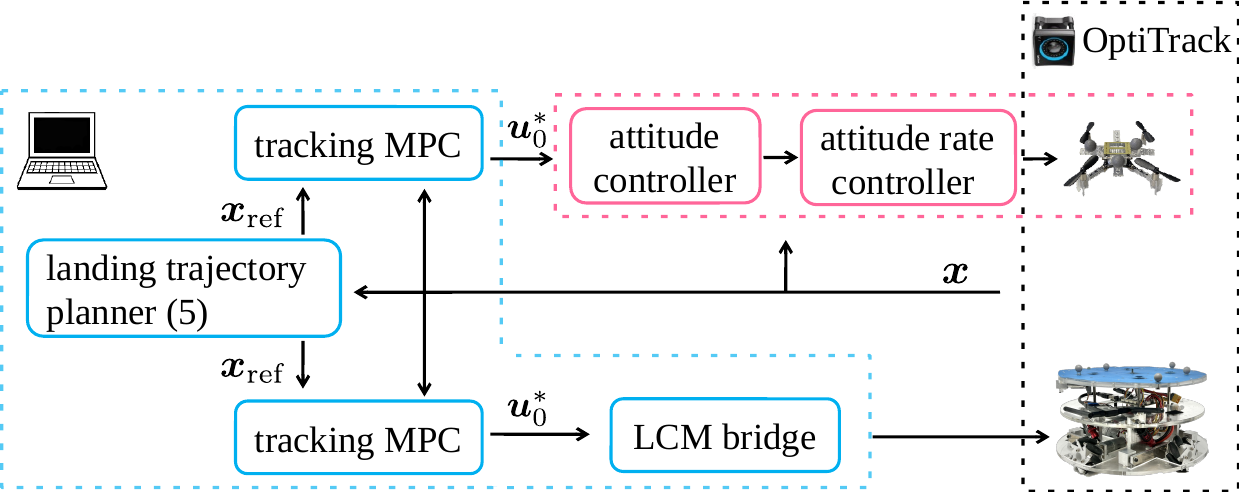}
  \caption{Control framework.}
  \label{fig:control_framework}
\end{figure}

Two nonlinear model predictive controllers (MPC) for trajectory tracking are designed - one for the quadrotor and one for the mobile robot. Each controller follows the generated landing trajectories $\bm{x}_\text{ref}$ obtained from \eqref{eq:final_optimization} for the quadrotor and for the mobile robot, separately and independently. In both tracking MPCs, the respective system models are incorporated to ensure accurate and robust tracking performance. The optimization problem to be solved in every time step $t$ can be formulated as

\begin{equation}
  \label{eq:mpc_formulation}
  \begin{aligned}
    \min_{\bm{u}^*_{(\cdot\mid t)}} & \ \sum_{k=t}^{t+H-1}  (\| \bm{x}_{(k\mid t)} - \bm{x}_{\text{ref},(k\mid t)} \|_{\bm{Q}}^2 + \| \bm{u}_{(k\mid t)} - \bm{u}_{\text{ref},(k\mid t)} \|_{\bm{R}}^2) \\
                                                  & \hspace{10mm} + \| \bm{x}_{(t+H\mid t)} - \bm{x}_{\text{ref},(t+H\mid t)} \|_{\bm{P}}^2  \\
    \text{s.t.} \ \
    & \bm{x}_{(t\mid t)} = \bm{x}(t), \\
    & \bm{x}_{(k+1\mid t)} = f_{\text{RK4}}(\bm{x}_{(k\mid t)}, \bm{u}_{(k\mid t)}) \ \forall k \in \{t, \ldots, t+H-1\}, \\
    & \bm{x}_{\text{min}} \leq \bm{x}_{(k\mid t)} \leq \bm{x}_{\text{max}} \ \forall k \in \{t, \ldots, H\}, \\
    & \bm{u}_{\text{min}} \leq \bm{u}_{(k\mid t)} \leq \bm{u}_{\text{max}} \ \forall k \in \{t, \ldots, t+H-1\},
  \end{aligned}
\end{equation}
where $\bm{Q}$, $\bm{R}$, and $\bm{P}$ are positive-definite weighting matrices and $H \in \mathbb{N}^+$ is the length of the optimization horizon. The function $f_\text{RK4}$ denotes the 4-th order explicit Runge-Kutta method utilized for discretizing the system dynamics. 
For the trajectory tracking, we employ a reduced-order model of the quadrotor and the kinematic model of the omnidirectional mobile robot, as detailed in~\cite{Llanes2024} and~\cite{Eschmann2023}. 
Given the highly dynamic nature of quadrotors, the tracking MPC must operate at a high update rate to ensure real-time adaptation and responsive tracking performance. In the experiments, the tracking MPC is executed at a rate of 100\,Hz for the quadrotor using the \textit{ACADOS} library with the sequential quadratic programming (SQP) method~\cite{Verschueren2021} and 10\,Hz for the mobile robot. 
At each control iteration, the first optimized control input $\bm{u}^*_{0}$ is applied to the quadrotor and the mobile robot, respectively. 
Notice that the input of the simplified quadrotor dynamics consists of the thrust $f_{\text{thrust}}$ and the attitude angles. Utilizing the part of the onboard controller within the \textit{Crazyflie} firmware, these inputs can be mapped to motor commands. The overall control framework is illustrated in Fig.~\ref{fig:control_framework}, where the section enclosed by the blue dotted line represents computations running on the ground station, implemented in \textit{ROS Humble}.

\subsection{Hardware Results}
\label{sec:real_world_result}

The quadrotor and the mobile robot are initially placed in separate locations within the experimental area. 
The cooperative landing sequence is initiated once the quadrotor achieves a stable hover. 
At this point, the motion capture system captures the initial states of both robots and relays this information to the landing trajectory planner.
Upon receiving the initial states, the landing trajectory planner generates cooperative landing trajectories for both the quadrotor and the mobile robot simultaneously. The robots immediately begin tracking these trajectories upon receiving.

\begin{figure}[htpb]
  \centering
  \includegraphics[width=\columnwidth]{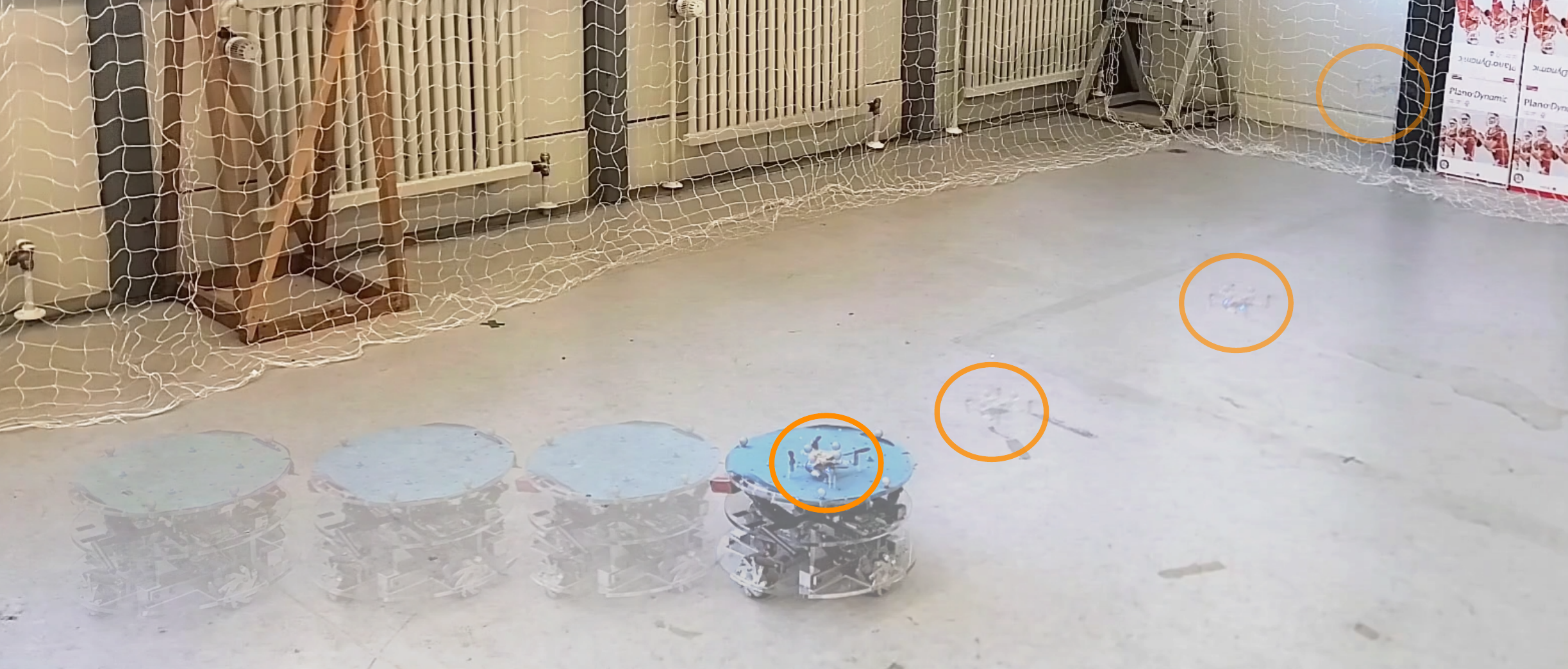}
  \caption{A composite image showing the real-world landing process of the quadrotor on the mobile robot. We use a color gradient from light to dark to indicate the direction of movement. For better visibility, we highlight the quadrotor with an orange circle.}
  \label{fig:landing_exp}
\end{figure}

\begin{figure}[htpb]
  \centering
  \includegraphics[width=0.8\columnwidth]{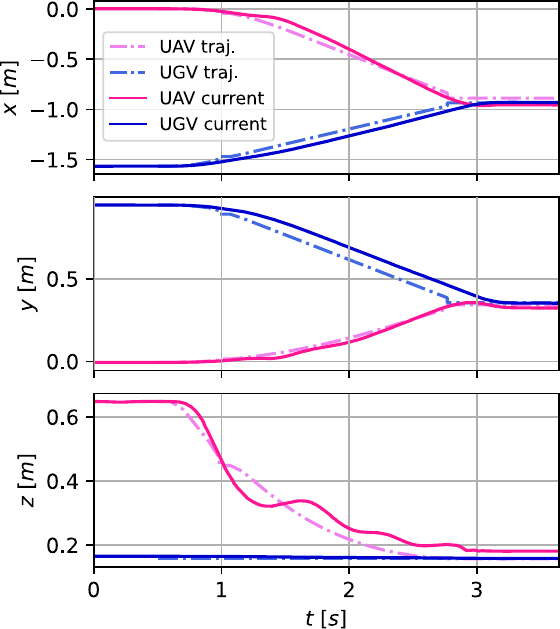}
  \caption{Output trajectories recorded by the motion capture system, labeled as \textit{UAV current} for the quadrotor and \textit{UGV current} for the mobile robot, and the reference trajectories generated by Eq.~\eqref{eq:final_optimization}, denoted as \textit{UAV traj.} and \textit{UGV traj.}}
  \label{fig:output_trajectories}
\end{figure}

\begin{figure}[htpb]
  \centering
  \includegraphics[width=0.8\columnwidth]{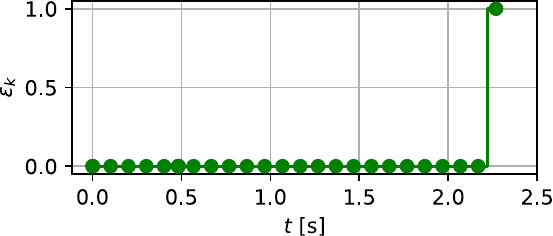}
  \caption{Values of $\varepsilon_k$ during the landing process.}
  \label{fig:epsilon_development}
\end{figure}

\begin{figure}[htpb!]
  \centering
  \includegraphics[width=0.8\columnwidth]{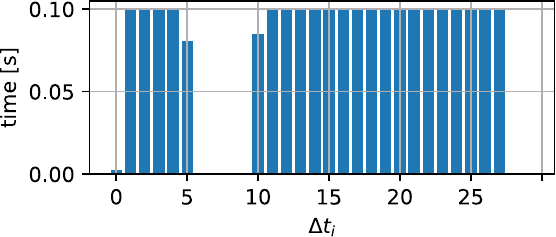}
  \caption{Results of $\Delta t_i$ for $i \in {\{0, \ldots, N}\}$, $\Delta t_\text{max}=0.1$\,s}
  \label{fig:tn_development}
\end{figure}

A landing trail is visualized in Figs.~\ref{fig:landing_exp} and~\ref{fig:output_trajectories}. In this instance, the quadrotor's initial position is $\bm{p}_{\text{q},0}=[0.0\ 0.0\ 0.65]^\top$\,m and the ground mobile robot starts at $\bm{p}_{\text{omni},0}=[-1.57\ 0.95\ 0.157]^\top$\,m. According to the initial distance, $N$ is autonomously set to $30$.
The quadrotor successfully lands on the mobile robot, with the entire landing process taking 2.268\,s. Figure~\ref{fig:epsilon_development} depicts the evolution of the landing index variable $\varepsilon_k$, which transitions from 0 to 1, signifying the completion of the landing maneuver. 
Figure~\ref{fig:tn_development} presents the optimization results for the time allocation variable $\Delta t_k$, providing insights into the temporal distribution of the trajectory waypoints.
Last but not least, the computation time required to solve the planning problem is 0.133\,s, which aligns with the observations from numerical simulations and demonstrates the real-time (re)planning ability of our framework.

\section{CONCLUSIONS}
\label{sec:conclusion}
In this work, we proposed a real-time optimization-based trajectory planning framework for cooperative landing between a quadrotor and a mobile robot, where the coordination of this heterogeneous multi-robot system is autonomously determined through a centralized decision-making process. Meanwhile, the time optimal control problem is considered to achieve an agile and efficient landing process.
The real-time planning capability of the framework is demonstrated by successful landing experiments. 
To the best knowledge of the authors, this paper's approach is currently the most speedy approach available that is as general as this, and can compete in calculation speed with heuristically simplified, more suboptimal approaches.
Future work will focus on extending the framework to consider additional constraints and uncertainties, exploiting the replanning capability, as well as integrating the framework with other cooperative tasks.

\section*{APPENDIX}
A video of the hardware experiments is available at \url{https://www.itm.uni-stuttgart.de/links/jschen/ITM_cooperative_landing}.

\section*{ACKNOWLEDGMENT}

This work was supported by the Deutsche Forschungsgemeinschaft (DFG, German Research Foundation) under Grants 433183605 and 501890093 (SPP 2353), and through Germany’s Excellence Strategy (Project PN4-4 Theoretical Guarantees for Predictive Control in Adaptive Multi-Agent Scenarios) under Grant EXC 2075-390740016.

\end{document}